\documentclass[conference]{IEEEtran}
\usepackage{url}
\usepackage{hyperref}
\usepackage[pdftex]{graphicx}
\usepackage{amsmath}
\usepackage{xcolor}
\usepackage{comment}
\hypersetup{draft}
\usepackage{balance}
\usepackage{parskip}
\usepackage[utf8]{inputenc}

\begin{document}
\title{
Towards the Internet of Robotic Things: Analysis, Architecture, Components and Challenges}
\date{}

\author{
\IEEEauthorblockN{Ilya Afanasyev\textsuperscript{1}, Manuel Mazzara\textsuperscript{1}, Subham Chakraborty\textsuperscript{1}, Nikita Zhuchkov\textsuperscript{1}, \\ Aizhan Maksatbek\textsuperscript{2}, Mohamad Kassab\textsuperscript{3}, Salvatore Distefano\textsuperscript{4}}
\IEEEauthorblockA{
\textsuperscript{1}Innopolis University, Innopolis, Russia \\
\textsuperscript{2}Yildiz Technical University, Istanbul, Turkey \\
\textsuperscript{3} Pennsylvania State University, PA, United States \\
\textsuperscript{4} University of Messina, Messina, Italy \\
\{i.afanasyev, m.mazzara, s.chakraborty, n.zhuchkov\}@innopolis.ru, \\ aizhanmaksatbek@gmail.com, muk36@psu.edu, sdistefano@unime.it}
}

\maketitle

\begin{abstract}
Internet of Things (IoT) and robotics cannot be considered two separate domains these days. Internet of Robotics Things (IoRT) is a concept that has been recently introduced to describe the integration of robotics technologies in IoT scenarios. As a consequence, these two research fields have started interacting, and thus linking research communities. In this paper we intend to make further steps in joining the two communities and broaden the discussion on the development of this interdisciplinary field. The paper provides an overview, analysis and challenges of possible solutions for the Internet of Robotic Things, discussing the issues of the IoRT architecture, the integration of smart spaces and robotic applications.
\end{abstract}

\section{Introduction}
Internet of Things (IoT) is a vivid and active research area \cite{IoT_Survey} and, at the same time, robotics is a solid and established field with numerous applications \cite{Royakkers2015, Preising1991, mohamed2018}. Although for some time the two directions continued intensively but separately, it is clear that modern scenarios require an integration of the two disciplines and a joint effort from the communities. With our work, we aim at developing this initiative. This paper surveys IoT and Robotics technologies together with their integration towards the realization of the \textit{Internet of Robotics Things} (IoRT) \cite{ray2016internet, simoens2018internet, batth2018internet, mahieu2019semantics}. We define several related concepts and we organize them in a coherent manner. This conceptual frame will be useful to the reader in order to identify the state-of-the-art literature and how to connect the dots into an holistic vision of the future synergy that has to unavoidably come between IoT and Robotics. The aim of this paper is to provide a better understanding of the IoRT and identify open issues that is worth investigating in future. In order to provide a comprehensive view of the area, the paper is organized as follows: Section \ref{sec:IoT} and \ref{sec:IoRT} introduce the reader to the ideas behind IoT and IoRT so that Section \ref{sec:IoRTArch} could discuss architectural aspects of the IoRT. Networking, multi-robots systems and Computing are respectively presented in Sections \ref{sec:networks}, \ref{sec:multi-robot} and \ref{sec:computing}, while Section \ref{sec:security} focuses on security concerns. Applications are discussed in Section \ref{sec:applications} and conclusive remarks and reflections reported in Section \ref{sec:conclusions}.


\section{IoT}
\label{sec:IoT}

The term “Internet of Things” (IoT) has recently become popular to emphasize the vision of a global infrastructure that connects physical objects/things, using the same Internet Protocol, allowing them to communicate and share information \cite{sula2013iot}. The term “IoT” was coined by Kevin Ashton in 1999 to refer to “uniquely identifiable objects/things and their virtual representations in an internet-like structure” \cite{uzelac2015comprehensive, han2011research}. According to analyst firm Gartner, 8.4 billion ‘things’ were connected to the internet in 2017; excluding the laptops, computers, tablets and mobile phones. This number is set to increase and reach 20.4 billion deployed IoT devices by 2020 \cite{Gartner17}.

IoT applications are already being leveraged in diverse domains such as medical services field, smart retail, customer service, smart homes, environmental monitoring and industrial internet. Now, due to their ubiquitous nature, the ``Internet of Robotic Things'' which binds together the sensors and the objects of robotic things is gaining popularity. However, a few challenges are there to maintain this trend. IoT is the enabler of Collaborative robots \cite{simoens2018internet} and several papers have already proposed IoRT-based architectural concepts \cite{ray2016internet, 5509469, 7354254}. All these research works are based on robotic systems to connect, share, and disseminate distributed computational resources, business activities, context information, and environmental data. Issues such as computational problems, optimization, and security are often still open challenges.


\section{Internet of Robotic Things}
\label{sec:IoRT}

\begin{figure}[!htbp]
    \centering
    \includegraphics[width=\linewidth]{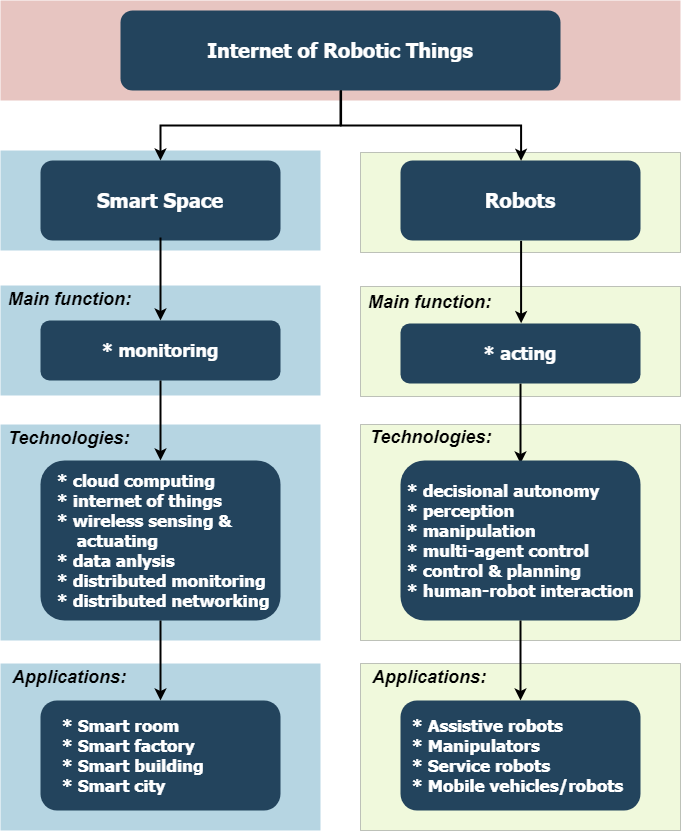}
    \caption{The Internet of Robotics Things block-scheme}
    \label{IoRT-scheme}
\end{figure}

Within the conceptual framework of IoT, the \textit{Internet of Robotic Things} concerns the integration of Smart Space capabilities and Autonomous agents (robots). This idea is pictorially represented in Figure \ref{IoRT-scheme}. In this context by Smart Space we mean applications like Smart Room, Smart Factory, Smart Building or Smart City \cite{mazzara2019reference}. The main function of these applications is the monitoring of states and processes in a defined controlling area. Other functions typically regards maintaining some desired environmental conditions, such as temperature and air humidity in the space by using sophisticated Heating Ventilation Air Conditioning (HVAC) system or by monitoring the states with simple sensors and actuators (e.g. turning on air conditioning or opening the window with simple drives, as well as timely switching on/off the heating). Managing power consumption is also one of the objectives sometime, for example by turning off the electric power by controlling human presence or launching of household appliances such as a washing machine, etc. when the daily electricity tariffs or power overloads are minimal.

Despite the availability of monitoring functions and simple actuators, Smart Space has no agents to perform indoor actions (moving objects, performing certain operations or services, etc.). Such agents are robots, such as assistive robots, manipulators, service robots, mobile vehicles/robots. The emergence of intelligent agents in the Smart Space completes the concept of the Internet of Robotic Things by integrating the functionalities of the Smart Space and Robots and expanding their possibilities. Modern roboticists frequently focus only on increasing the level of robot autonomy, enhancing the requirements for perception with robot sensors and onboard data processing that should allow robots to perform tasks independently. Very often they ignore the fact that indoor environment, where robots execute certain operations, can be filled with various sensors (RFID, occupancy sensors, surveillance cameras, magnetic sensors, IR/sonic beacons, etc.) and computing resources (smartphones, routers, computers, servers, etc.). Thus, both robots, whose functionalities are increased by the smart environment resources, and the Smart Space benefit from such integration, in which besides monitoring functions and performing simple actions with simple mechanisms, agents (robots) appear to perform complex operations inside the Smart Space. As a result of this integration, robots can receive tasks from the Smart Space (for example, Smart Building or Smart Factory), which also monitors the progress in task execution and gives hints to robots from the Smart Environment sensor network, for example, for optimal navigation, obstacle/collision avoidance or effective human-robot interaction. Thus, the Internet of Robotic Things is a more advanced level of the Internet of Things, allowing to integrate such modern technologies as cloud computing, wireless sensing and actuating, data analysis, distributed monitoring and networking from the Smart Space, as well as decisional autonomy, perception, manipulation, multi-agent control, control and planning and human-robot interaction - from the robot side (Figure \ref{IoRT-scheme}).

\section{IoRT Architecture}
\label{sec:IoRTArch}

\begin{figure*}[!htbp]
	\centering
	\includegraphics[width=\linewidth]{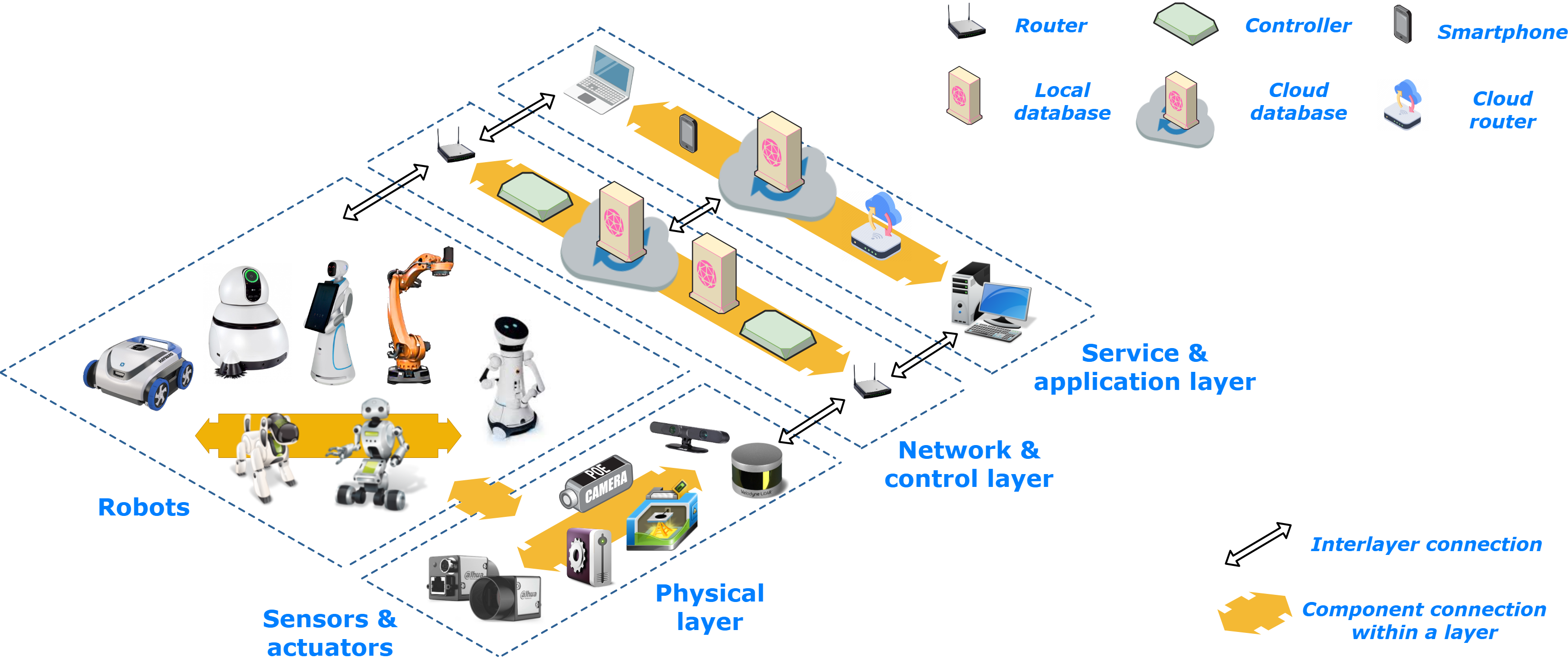}
	\caption{The Internet of Robotic Things system reference architecture}
	\label{IoRT-architecture}
\end{figure*}

In this section, we present the three-level reference architecture of the Internet of Robotic Things (Figure \ref{IoRT-architecture}), which is one of the contributions of this article. Figure 2 demonstrates the overall system architecture consisting of three main layers: 1. \textit{Physical}, 2. \textit{Network and Control}, and 3. \textit{Service and Applications}.

The \textit{Physical layer} is represented by various Robots, Sensors and Actuators. The Robots are intelligent agents that can communicate with each other and establish a multi-robot system (for details, see Section \ref{sec:multi-robot}) to achieve a common goal through distributed actions. Sensors at the Physical layer are devices for monitoring vital environmental parameters, as well as perceptions of the Smart Space to observe what processes occur in the room: which agents (people or robots) are present, which objects move, what actions happen, etc. On the Physical level there are also simple drives, switches and actuators that can perform simple actions (turn on/off heating, lighting, air conditioning, etc.), and even machine tools or 3D printing devices that can produce details for the Smart Space objectives. In special cases, Robots can apply Sensors and Actuators directly for their activities (for example, as landmarks for navigation or for calibration and adjustment), optimizing processes in the Smart Environment. However, the main way to integrate Robots with Sensors and Actuators in the Smart Space network occurs at the Network and Control layer, where different components can utilize common/different protocols to communicate and control processes in the Smart Environment. The \textit{Network and Control} layer can include various routers, controllers, local and cloud data storages (servers), as well as communication and control protocols (for more details, see Section \ref{sec:networks}). For preliminary processing and storage of data from Sensors, Actuators and Robots, both local storage (in each Smart room or the Smart building as a whole) and remote storage (in the clouds) can be exploited. At the \textit{Service and Application} layer, the implementation and execution of standard and user programs for monitoring, processing and controlling both environmental parameters and agents (Sensors, Actuators and Robots) in the Smart Space are performed in accordance with the objectives of the integrated Internet of Robotic Things. In addition to state-of-the-art algorithms for processing sensory information at this layer, artificial intelligence (AI) and machine learning (ML) algorithms can be utilized to optimize the IoRT performance, using the fact that modern databases provide low latency in data transfer.

\section{Networks}
\label{sec:networks}

\begin{figure*}[!htbp]
	\centering
	\includegraphics[width=\linewidth]{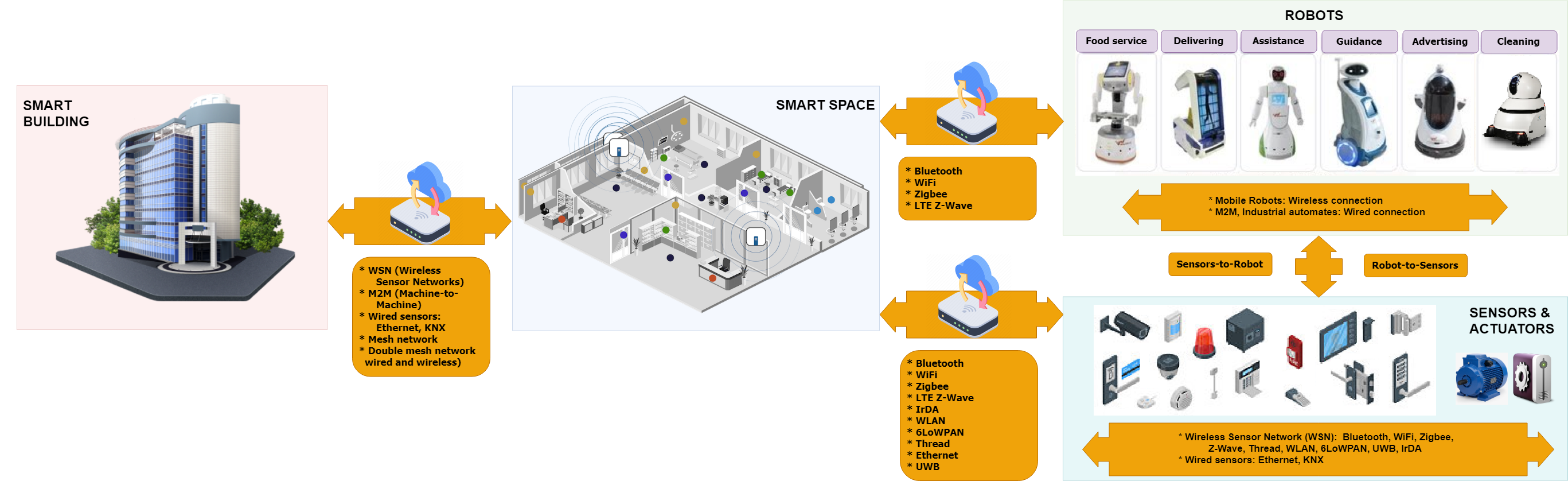}
	\caption{The functional diagram of network interaction between Robots, Sensors and Actuators, Smart Space and Smart Building}
	\label{Robot-Networks}
\end{figure*}

The Internet of Robotic Things technology implicates the implementation of a fully distributed system that extends to the various levels - Smart Building, Smart Space, Robots, Sensors and Actuators, in which both wired and wireless networks of different ranks (not only peer-to-peer interactions) can be presented. They should be equipped with controllers that collect, analyze and transfer data through gateways to other networks, computing devices, clouds or autonomous agents (e.g. robots/intelligent machines for onboard decision-making). The functional diagram of the network interaction between Robots, Sensors and Actuators, Smart Space and Smart Building is presented in Figure \ref{Robot-Networks}.

It is known that network protocols for Smart Buildings are divided into smart device networks and traditional networks for high-speed data transfer \cite{mazzara2019reference}. Therefore it is logical to use the protocols already installed in the Smart Building infrastructure (such as wireless sensor networks (WSN) and machine-to-machine communications (M2M)) for IoRT components and agents. The mesh network is often a suitable choice of network topology for wireless communication of robots and devices inside Smart Space due to the indoor obstacles in the Smart Environment (walls, furniture, interior items, etc.). Double mesh with wired and wireless networks is appropriate for those Smart Buildings / Smart Factories where was already installed a wired automation system \cite{mazzara2019reference}. At present there are many wired and wireless communication networks and their protocols used to exchange data with devices, servers and clouds, components, sensors and robots, which are described in \cite{DEKRA2017, mazzara2019reference} and can be also used in IoRT (they are shown schematically in Fig. \ref{Robot-Networks}).

\section{Multi-robot systems}
\label{sec:multi-robot}
An interesting application scenario for the Internet of Robotics Things is the multi-robot systems (MRS) one.
Multi-robot systems are robotic systems composed of multiple (robotic) agents operating in the same environment, which could implement simple sensing/actuation activities or even complex operations \cite{MRS}.
Various rather interchangeable terms are used in this area to characterize multi-robot systems, such as group robotics, swarm robotics, collective robotics, cooperative robotics.
These terms can differ in the range or the number of robots involved, but they can be framed altogether under the big and generic umbrella term of multi-robot systems.
Robots are grouped into multiple-swarm-collective-cooperative sets to basically improve the overall system performance (by increasing the throughput), extend and enable new functionalities of the robotic system, perform distributed activities even remotely and improve fault tolerance through redundancy.
From  the coexistence of multiple robots potential drawbacks that may arise concern
interference between robots, overhead, cost and robustness, control and coordination.

A taxonomic classification of multi-robot systems \cite{MRSTAXO} identifies two groups of dimensions: Coordination Dimensions, referring to the type of coordination that is achieved in the MRS, and System Dimensions, including the system features that govern the coordination.
The former group includes aspects related to cooperation (cooperative, competitive), knowledge (aware vs unaware), coordination (strongly, weakly, no coordination) and organization (strongly centralized, weakly centralized, distributed).
The latter features concern communication mechanisms (networking, topology, stigmergy), team composition (heterogeneous vs homogeneous), system architecture (reactive vs deliberative) and team size (small, medium, big).
IoT and IoRT could perfectly fit with the networking requirements framed into the communication system dimension of MRS, coming as a natural solution for all such issues.

\section{Computing}
\label{sec:computing}

The Internet of Robotics Things (IoRT) usually generates and process a large quantity of data for properly operating. At the same time robots, even the ones of large size, have a small amount of place and energy for computing and data storage. The same happens to IoT devices, whose main functionalities are generating data, control state and pass commands. Therefore, it is necessary to add external calculation and storage resources to the IoRT network. These resources are typically provided by Cloud computing and cloud-related architectures.

\subsection{Cloud computing}
Cloud internet of robotics things has recently emerged as a collaborative technology between cloud computing, IoT devices and service robotics enabled through progress in wireless networking, large scale storage and communication technologies, and the ubiquitous presence of Internet resources over recent years \cite{saha2018}. Before bringing IoRT into the cloud we should consider next points \cite{mouradian2018robots}:
\begin{itemize}
    \item Latency-sensitivity requirement of IoRT applications needs to be considered as a long latency will generate high risks for real-time applications;
\item Limited computation and autonomy capabilities of IoT devices, with regards to VMs, when they are virtualized and used. 
\item Wide diversity and heterogeneity of IoT and robotics devices are often difficult to aggregate and integrate into the one system.
\end{itemize}
All risks described above could be covered by fog and edge computing.
\subsection{Edge computing}
Edge computing or Multi-Access Edge Computing (MEC) generally handles the processing of data where data is created around the network in substitute of centralized data-processing warehouse. For the entry point, edge devices are used that enables the entry into the core networks. Here, computation is largely or completely performed on distributed device nodes known as smart devices or edge devices as opposed to primarily taking place in a centralized cloud environment.
Edge computing could provide advantages in the next cases:
\begin{itemize}
\item Autonomous connected vehicles. 
Self-driving cars should be able to work and learn without constant connection to the cloud to process data. But vehicle communicate with the infrastructure, it may talk to the other vehicles around it, but most of its onboard processing;
\item Predictive maintenance.
Edge computing can help detect machines that are in danger of breaking, and find the right fix before they do. As the alert should be generated with the smallest possible latency it's should be done closer to that machine.
\item Supporting of the fog architecture.
Fog computing refers to a distributed computing model, in which edge (peripheral) devices are used as terminals for computing.
\end{itemize}
\subsection{Fog computing}
Fog computing is a architecture that defines how edge computing should be organized, and it facilitates the operation of compute, storage and networking services between end devices and cloud computing resources. 

Fogging extends the concept of cloud computing to the network edge, making it ideal for Internet of things and other applications that require real-time interactions. Fog networking mainly utilizes the local computer resources rather than accessing remote computer resources causing a decrease of latency issues and performance further making it more powerful and efficient \cite{mohamed2018}.

Fog computing advantages \cite{Kehoe2015survey}:
\begin{itemize}
    \item Reduces amount of data sent to cloud;
    \item Minimize network latency;
    \item Supports mobility;
    \item Conserves network bandwidth;
    \item Improves system response time.
\end{itemize}

\section{Security}
\label{sec:security}

Security is a complex and challenging issue in this area, related both to the security of the Internet of Things and to the security of robots connection. Main cybersecurity problems in robotics arise due to some of the following reasons listed below:
\begin{itemize}
    \item Insecure communication between users and robots lead to cyber-attacks. Hackers can easily hack into insecure communication link in no time.
    Hackers can easily hack into insecure communication link in no time.
    \item Authentication issues. Failure in guarding against unauthorized access can easily allow hackers to enter the robot systems and use their functions from remote locations without using any valid username and password. 
    \item Lack of proper encryption at vendors' side that can expose sensitive data to potential hackers.
    \item Most of the robot features are programmable and affordable. If the default robot configuration is weak to hacking, intruders can easily get access to the programmable features and change them.
\end{itemize}


The problem of cybersecurity in robotics is outlined in \cite{dragoni2016internet}, which discusses existing bugs and vulnerabilities that admit robots to be hacked remotely, as well as applications that require security and privacy to be implemented in the field of robotics.
~
~
Cybersecurity is also relevant for robots and automation systems that rely on data and software code from the network to maintain their functionality. This problem also affects big data processing and cloud computing due to access to libraries, datasets, maps, etc. and cloud operations that can also be associated with access to parallel grid computing using on-demand statistical analysis, and therefore they should be included in the security umbrella \cite{Kehoe2015survey}.


IoT-based applications for robotics require solving some problems, developing methodologies and choosing architectural solutions \cite{grieco2014}. Cybersecurity is also related to data transfer and processing with communication protocols, therefore such communications must be encrypted, although in most cases it is not occured \cite{hu2012cloud}.
While having human-robot interaction there is a potential danger of interfering in such communications, leading to changes in commands to robots. If there is no encryption or authentication mechanism that controls such an interface, the system is prone to man-in-the-middle attacks.

Let's consider household robots. In the next decade, it is expected that every house will have robots, e.g. home assistants in daily chores. They may contain microphones, cameras and sensors that will collect datasets, including personal information about house and even people's health status. Insufficient care about protection of this confidential information may result in gaining a control under such service robot and accessing confidential data by an unauthorized entity.

Cybersecurity of IoT systems using cloud computing is another challenging problem, since IoT devices can be connected through a cloud, providing cloud communication and data collection. In this case, protection against DDoS attacks becomes an important element of the system security \cite{DeDonno2019}.


\section{Applications and Services}
\label{sec:applications}

Integration of robotics into smart spaces can be utilized in various areas of our life: home automation, health, transportation, logistics.
As a smart space concept, we can introduce Smart and Software Defined Buildings (SSDB) \cite{mazzara2019reference}. These are “programmable” buildings where sensing, based on hardware and software, is integrated to perform various functions such as presence monitoring, activity and identity recognition, and detection of user’s emotional state. The sensing IoT functions are implemented using various hardware components.
Occupancy detectors are special circuits that detect individual’s presence with its motion sensor. Positioning and Tracking sensors are used in wearable devices to track individual’s movements. The obtained raw data is processed using one of the computation methods and used as a knowledge for robots. 
Because robots have limited memory and sensing capabilities, SSDB can be used as a distributed robotic sensing system.	 It will increase robot’s autonomy and capability.  As a result, it will be possible to use a robot just as an actuator for various services \cite{chamberlain2016}.
Let's discuss three application examples of IoRT presented in \cite{mahieu2019semantics} that include smart spaces and possible services: \textit{Smart Home}, \textit{Smart Office}, and \textit{Smart Nursing House}.

\textbf{Smart Home}.
The house is equipped with appropriate sensors for robot’s better context-awareness. A service robot can help with a household chore. Nutrition monitoring service gives meal advice and reminds fixed meal routines, the smart space tracks the meal information, after the robot actuates appropriate advice to the individual. Homework assistance service helps children with homework and also capable of choosing the appropriate place in house for doing a homework. Therapy monitoring service is helpful for individuals with various diagnoses such as a diabetes, it monitors sugar level in an organism and reminds about keeping a glucose level.

\textbf{Smart Office}.
Visitor Reception Service can assist meeting organizers. Concierge robots can greet a guest in a native language and show the meeting room. The robot is context-aware and is supplied with guest’s scanned QR-code and previous visits knowledge. Also, Visitor Reception can show optimal lifting method, if the guest is in a wheelchair.

\textbf{Smart Nursing House}.
Nursing House (NH) gives special person-centric care for patients with diverse diagnoses. Some patients need 24/7 care, but it requires a big budget. IoRT services can be integrated to reduce spendings on routine tasks \cite{mahieu2019semantics}. Every morning activity announcement service can announce planned activities and news. The service robot is adoptable to individual’s conversation preferences such as voice tone and greeting style. Behavioral Disturbance (BD) management is helpful for people with dementia. IoRT can partially replace the care staff and if the NH care staff help is needed the service can call for a help. The Smart space can detect a person showing behavioral disturbance and a service robot can assist in managing BD-people for a while. Such as detecting a disturbed elderly and calming an elderly by playing a favorite music or reminding past happy memories. Also, smart house can detect a wandering elderly and humanoid robot can lead an elderly to the own room. Visitor Information Announcement Service helps patients with mild concussion diagnose to regulate a room light and interior sound conditions. Various sensing devices are installed in the patient's room, if the abnormal sound level or light level detected, the smart NH robot is requested to notify the patient and visitors to regulate the conditions.


\section{Discussion and Conclusions}
\label{sec:conclusions}

The Internet of Robotics Things (IoRT) is a freshly introduced concept aiming at describing the integration of robotics technologies in IoT scenarios. Recently, the IoT and robotics research communities have started vividly interacting. This paper is an attempt to integrate further the two communities and develop this interdisciplinary field. 

The paper represents an comprehensive overview of the concepts and challenges in the IoRT and proposes the IoRT architecture. However, several complementary aspects are left out for future discussion. First, requirements engineering and formal process modeling \cite{Mazzara2010, YanMCU07}, a broad research area that can also be explored in this specific application domain. Second,
process reconfiguration \cite{Abouzaid:2013}, which especially applies to multi-robot systems as described in the related section of this work.
Third, although security has been discussed in this and previous papers \cite{dragoni2016internet, DonnoDGM16}, the open challenges are far from being solved here.

To address IoRT security problems, the certain vulnerability reduction techniques known in cybersecurity and have proven to be effective in resolving IoT security issues  over the years can be used. Among them are security analysis (by collecting, comparing and analyzing data from several sources and assisting IoT security providers in identifying potential threats); Implementing a public key infrastructure (i.e., a set of policies for software/hardware and procedures necessary to create, manage, and distribute digital certificates); Providing device authentication; Comprehensive device authentication.

Since the IoRT system works on the connection between the connected devices, when the connection is broken, a failure occurs that can disrupt the work and make it impossible for the robots to achieve their goals. Thus, securing communications is another open problem where network protection provides IoRT system security.


\balance
\bibliography{main}
\bibliographystyle{ieeetr}
\end{document}